\documentclass[letterpaper]{article} 
\usepackage{aaai2026}  
\usepackage{times}  
\usepackage{helvet}  
\usepackage{courier}  
\usepackage[hyphens]{url}  
\usepackage{graphicx} 
\usepackage{natbib}  
\usepackage{caption} 
\usepackage{amsmath}
\usepackage{amssymb}
\title{xLightGCN: Explainable Graph Neural Network Based Recommendation Systems with Most Influential Peers}

\author{
    Calder Johnson
    Robin Cohen
}
\affiliations{
    Computer Science, University of Waterloo, Canada\\
    cbjohnson@uwaterloo.ca
    rcohen@uwaterloo.ca
}

\begin{document}

\maketitle

\begin{abstract}
Recommendation systems, systems that predict a user's preference for items in order to curate that user's feed, are a ubiquitous and central feature in online social networks. While they are a central piece of our modern digital infrastructure, many modern approaches are opaque black boxes, making them difficult to audit, understand, and assess the trustworthiness of. In this paper we present an exact, graph-based approach for generating explanations for decisions made by recommendation systems. We argue that in the risk-sensitive sociotechnical context of user-facing AI such as recommendation systems, explainability is a key pillar for establishing the trustworthiness of models and validating their effectiveness. We propose a technique built on LightGCN-like architectures, an influential and popular class of recommendation models. We describe our approach and argue that it provides an appropriate assurance mechanism for these AI systems leading to specific advantages, including interpretable explanations and a theoretically guaranteed measure of explanation fidelity. We conclude by discussing limitations and directions for future research.
\end{abstract}

\begin{links}
\link{Code}{https://github.com/CalderJohnson/xLightGCN}
\link{Dataset}{https://snap.stanford.edu/data/loc-gowalla.html}
\end{links}

\section{Introduction}

Recommendation systems have become a central component of modern digital infrastructure. Online platforms use them to determine which products, social media posts, advertisements, etc., are shown to individual users. By predicting user preferences and recommending only a small subset of potentially relevant items from the full space of possibilities, recommendation systems can substantially improve the usability and engagement of online platforms. At the same time, the increasing role of automated recommendation in determining what users encounter creates an important challenge for trustworthy AI: the decisions of these systems must be understandable and auditable by the developers and organizations responsible for them.

Graph-based recommendation systems are particularly effective at exploiting the relational structure inherent in user behavior. A recommendation dataset can naturally be represented as a bipartite graph in which users and items are vertices and interactions such as purchases, views, or clicks form edges. This representation allows information to propagate through chains of interactions, enabling a model to exploit higher-order collaborative relationships that are difficult to represent using conventional recommendation approaches. LightGCN~\citep{he2020lightgcn} is an influential architecture in this paradigm, employing a simplified GCN (graph convolutional network) with no feature transformations and nonlinear activation functions while achieving strong empirical performance.

Given a set of users, a set of items, and the graph of past interactions between them, LightGCN predicts the affinity between a given user $u$ and item $i$ by producing initial node embeddings $H^0$ and running them through $L$ convolutional layers. At each layer $l$, the following operation is performed, where $A$ is the graph's adjacency matrix, and $D$ is its degree matrix:
\begin{align}
\hat{A} &= D^{-1/2}AD^{-1/2} \\
H^{l+1} &= \hat{A}H^{l}
\end{align}
such that the embeddings at layer $l$ are a degree-weighted linear combination of the embeddings of nodes $l$ hops away in the graph. Next, embeddings at each layer are combined in a weighted sum ($\alpha_0, \ldots, \alpha_L$ being hyperparameters), and the affinity score $s$ between a user and an item can be measured by an inner product:
\begin{align}
H_{out} &= \alpha_LH_L + \alpha_{L-1}H_{L-1} + \ldots + \alpha_0H_0 \\
s(u, i) &= \langle H_{out}^u, H_{out}^i \rangle
\end{align}

LightGCN's simple, entirely linear structure and groundbreaking empirical results made it a highly influential model, inspiring a wider class of linear GCN architectures such as LGKAT~\citep{hassanzadeh2025novel} which extends LightGCN using attention layers and DualLightGCN~\citep{huang2023duallightgcn} which splits the architecture into two networks that the methods discussed in this paper could be extensible to in future research.

Despite their power, the opacity of these models presents an important assurance problem. A recommendation depends on complex chains of interactions involving other users and items that provide a rich collaborative signal to the model but are intractable for humans to inspect at the scale of a real interaction graph. Moreover, interaction histories can encode highly personal, detailed information about user behavior and preferences. A misaligned recommendation model operating on sensitive user data whose decision making process is opaque presents significant risk from a sociotechnical perspective. For example, without an appropriate assurance framework in place, such a system could promote echo chambers, give identity-biased recommendations, or hurt brand image by targeting ads to the wrong demographics.

We view explainability as a mechanism for producing evidence that can support the assessment of AI-system trustworthiness. By exposing the evidence underlying a decision, developers and auditors can inspect model behavior and identify potentially problematic dependencies, allowing them to make an informed assessment of trustworthiness. The ability to identify which relationships contributed to a recommendation is therefore key to the transparency, risk assessment, and trustworthiness judgement of user-facing AI systems.

Existing explanation methods for GNNs often produce approximate or computationally expensive explanations. GNNExplainer~\citep{ying2019gnnexplainer}, for example, identifies informative subsets of a graph through an optimization procedure that masks nodes and edges while attempting to preserve the model's prediction. SubgraphX~\citep{yuan2021subgraphx} instead uses a search procedure based on Monte Carlo Tree Search and Shapley values to identify influential subgraphs. Recommendation-specific approaches have also produced interpretable reasoning chains through knowledge graphs~\citep{wang2019kgrecommendation,wang2018ripplenet}. These methods demonstrate the feasibility of generating human-interpretable explanations, but they generally either rely on additional optimization or search procedures for individual predictions. This motivates the question of whether the linear structure of LightGCN can be exploited to obtain explanations that are both computationally efficient and directly grounded in the model's actual computation.

In this work, we propose xLightGCN, an explanation framework that identifies the most influential peers underlying individual LightGCN recommendations. Using the linear structure of LightGCN, we precisely isolate the compact explanatory subgraph that explains most of the model's prediction in an efficient manner. In addition, the component of the prediction explained by the subgraph can be compared to the complete prediction to quantify the contribution left unexplained by the extracted subgraph, providing a theoretically grounded measure of explanation fidelity.

\section{Methods}

We formulate explanation of LightGCN recommendations as a node contribution problem. Given a user $u$, target item $i$, and propagation depth $L$, xLightGCN identifies the nodes whose initial embeddings make the largest contributions to the resulting recommendation score and constructs a compact subgraph around them. The subgraph reveals the most critical relationships in the network that influenced a recommendation decision, providing insight into how the model came to a particular prediction, improving its transparency. The insights gained from this improved transparency could be leveraged by developers to assess the validity and trustworthiness of the system.

Let $e_v^{(0)}$ denote the initial embedding of node $v$. Since LightGCN propagation is linear, the final embedding of an endpoint can be expressed as a linear combination of the initial embeddings of nodes within its $L$-hop receptive field. Thus,
\begin{align}
e_u &= \sum_{v \in V_u} c_{u,v} e_v^{(0)}, \\
e_i &= \sum_{v \in V_i} c_{i,v} e_v^{(0)},
\end{align}
where $V_u$ and $V_i$ are the $L$-hop neighborhoods of $u$ and $i$, respectively, and $c_{u,v}$ denotes the coefficient of node $v$'s initial embedding in the corresponding final embedding.

The algorithm first constructs the union of these two receptive fields, including all edges between nodes contained within them. For each node in the user-side receptive field, we compute its alignment with the target item's final embedding,
\begin{equation}
p_{u,v} = (e_v^{(0)})^\top e_i
\end{equation}
and define its recommendation-level contribution as
\begin{equation}
a_{u,v} = c_{u,v} p_{u,v}.
\end{equation}
Analogously, for nodes in the item-side receptive field,
\begin{align}
p_{i,v} &= (e_v^{(0)})^\top e_u, \\
a_{i,v} &= c_{i,v} p_{i,v}.
\end{align}

The magnitude $|a_{i,v}|$ therefore measures the extent to which a node influences the recommendation score, while its sign indicates whether the node contributes positively or negatively. xLightGCN selects the $k$ highest-magnitude contributors from each receptive field. For each selected node, a representative path connecting it to the corresponding endpoint is retained, and the union of these paths forms the final explanatory subgraph.

\begin{figure*}[t]
    \centering
    \includegraphics[width=0.8\textwidth]{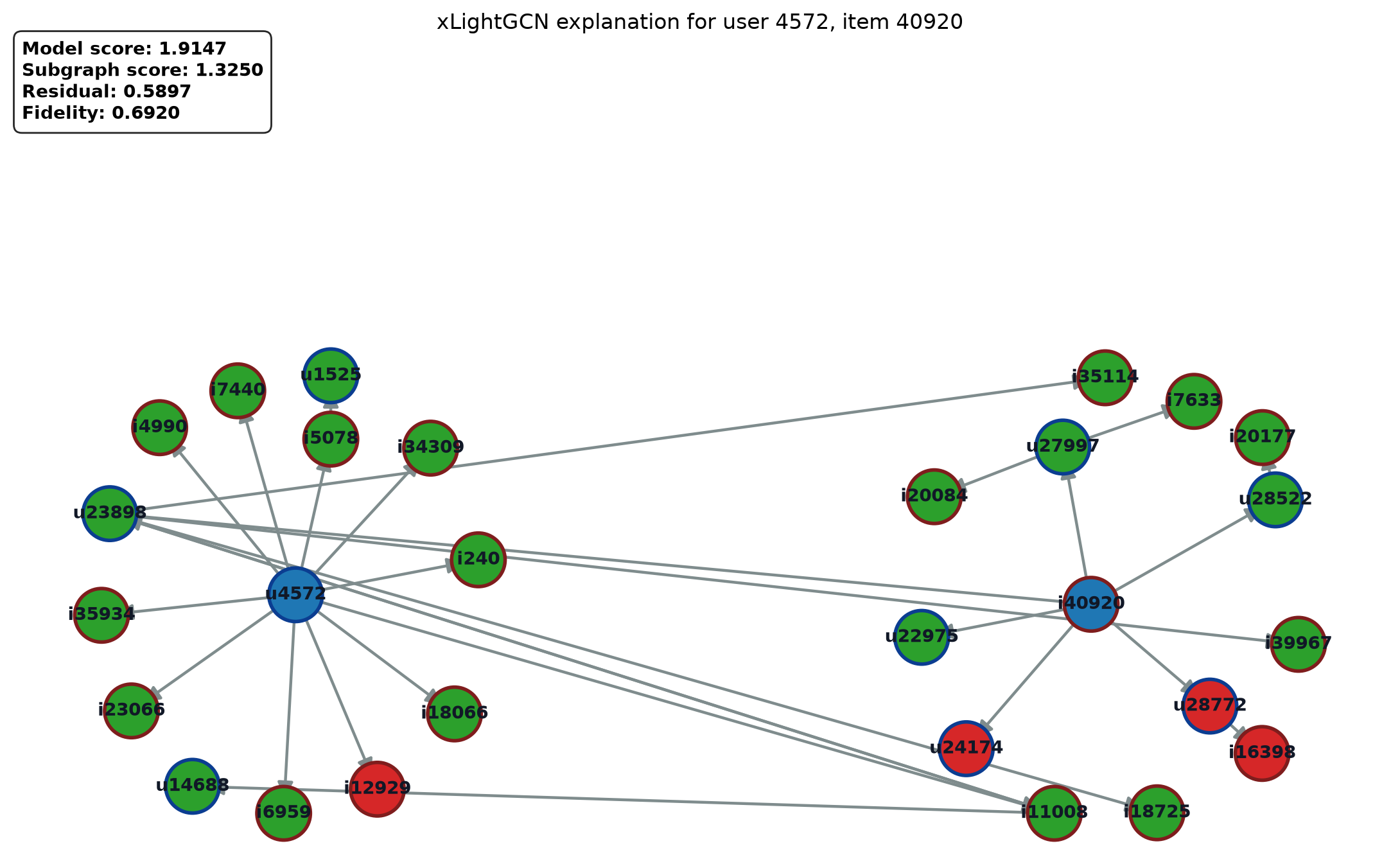}
    \caption{Example xLightGCN explanatory subgraph, showing the target user and item taken from the Gowalla dataset (where users are recommended interesting locations). Shown are the user and item in question (blue), selected positive (green) and negative (red) contributing nodes, and representative paths (intermediate nodes would be in grey). In this example, k=15 and fidelity shows that \textasciitilde 69\% of the score is explained by the subgraph. The linkages between the user and item subgraphs demonstrate the collaborative filtering relationships xLightGCN uncovers.}
    \label{fig:explanatory-subgraph}
\end{figure*}

xLightGCN is derived directly from the LightGCN computation and therefore does not require a separate surrogate model or per-instance model optimization. The $L$-hop receptive fields are constructed by bounded-depth frontier expansion in $O(m)$ time, where $m$ is the number of edges traversed in the resulting local subgraph; equivalently, the receptive field contains at most $O(d^L)$ nodes when (d) is the maximum node degree. Because LightGCN propagation is linear, the contribution coefficient $c_v$ for every node can be obtained by propagating a scalar coefficient vector through the same $L$ local message-passing steps, requiring $O(Lm)$ time. Given these coefficients, all recommendation-level node contributions can be computed efficiently by a matrix-vector product between the $n$ initial node embeddings and the opposing endpoint embedding, requiring $O(nD)$ time for embedding dimension $D$, followed by an element-wise multiplication with the coefficient vector. Selecting the $k$ largest-magnitude contributions can then be performed in $O(n\log k)$ time using a heap-based top-$k$ selection, and constructing representative paths requires $O(kL)$ time. Thus, the overall explanation procedure requires
\begin{equation}
O(Lm+nD+n\log k+kL)
\end{equation}
time on the local receptive field. For fixed propagation depth $L$ (usually only 2--4 in practice), embedding dimension $D$, and explanation size $k$, this scales with the size of the local receptive field rather than the full interaction graph. In contrast, surrogate-based approaches such as SubgraphX require repeated model evaluations during their per-instance search procedure, including evaluations of candidate subgraphs for Monte Carlo Tree Search and Shapley-value estimation. xLightGCN instead performs a single direct decomposition of the existing LightGCN computation, avoiding the repeated forward evaluations and optimization/search overhead characteristic of such surrogate-based approaches, allowing for highly efficient explanation generation.

An important property of xLightGCN is that it provides not only an explanation, but also a measure of how completely that explanation accounts for the model's prediction. Let $s(u,i) = e_u^\top e_i$ denote the original recommendation score. Given selected node sets $K_u$ and $K_i$, we construct truncated embeddings
\begin{align}
\tilde{e}_u &= \sum_{v \in K_u} c_{u,v} e_v^{(0)}, \\
\tilde{e}_i &= \sum_{v \in K_i} c_{i,v} e_v^{(0)},
\end{align}
and obtain the explanation-only score
\begin{equation}
\tilde{s}(u,i) = \tilde{e}_u^\top \tilde{e}_i.
\end{equation}
The difference
\begin{equation}
R(u,i) = s(u,i) - \tilde{s}(u,i)
\end{equation}
is the unexplained residual: the portion of the original prediction not captured by the selected explanatory nodes. We define explanation fidelity as
\begin{equation}
F(u,i) = 1 - \frac{|R(u,i)|}{|s(u,i)| + \epsilon}.
\end{equation}
where a small constant is added to the denominator for numerical stability. Higher fidelity indicates that the explanatory subgraph more completely reproduces the original model prediction. Fidelity score falls into the range $(-\infty, 1]$. Values approaching 1 indicate that the selected subgraph accurately reconstructs the original prediction. Values near 0 indicate substantial reconstruction error, while negative values indicate that the reconstruction error exceeds the magnitude of the original score.

This provides xLightGCN with an assurance-oriented property beyond qualitative interpretability. The explanatory subgraph provides a compact and inspectable representation of the model's computational evidence, supporting transparency and accountability, while $F(u,i)$ provides a simple quantitative measure of how much of the prediction that evidence accounts for. Low-fidelity explanations indicate that the selected evidence provides limited coverage of the model's prediction and may therefore warrant additional auditing or investigation. In this way, xLightGCN provides measurable evidence for assessing model behaviour and understanding its trustworthiness.

\section{Experiments}

\begin{figure}[t]
    \centering
    \includegraphics[width=0.9\columnwidth]{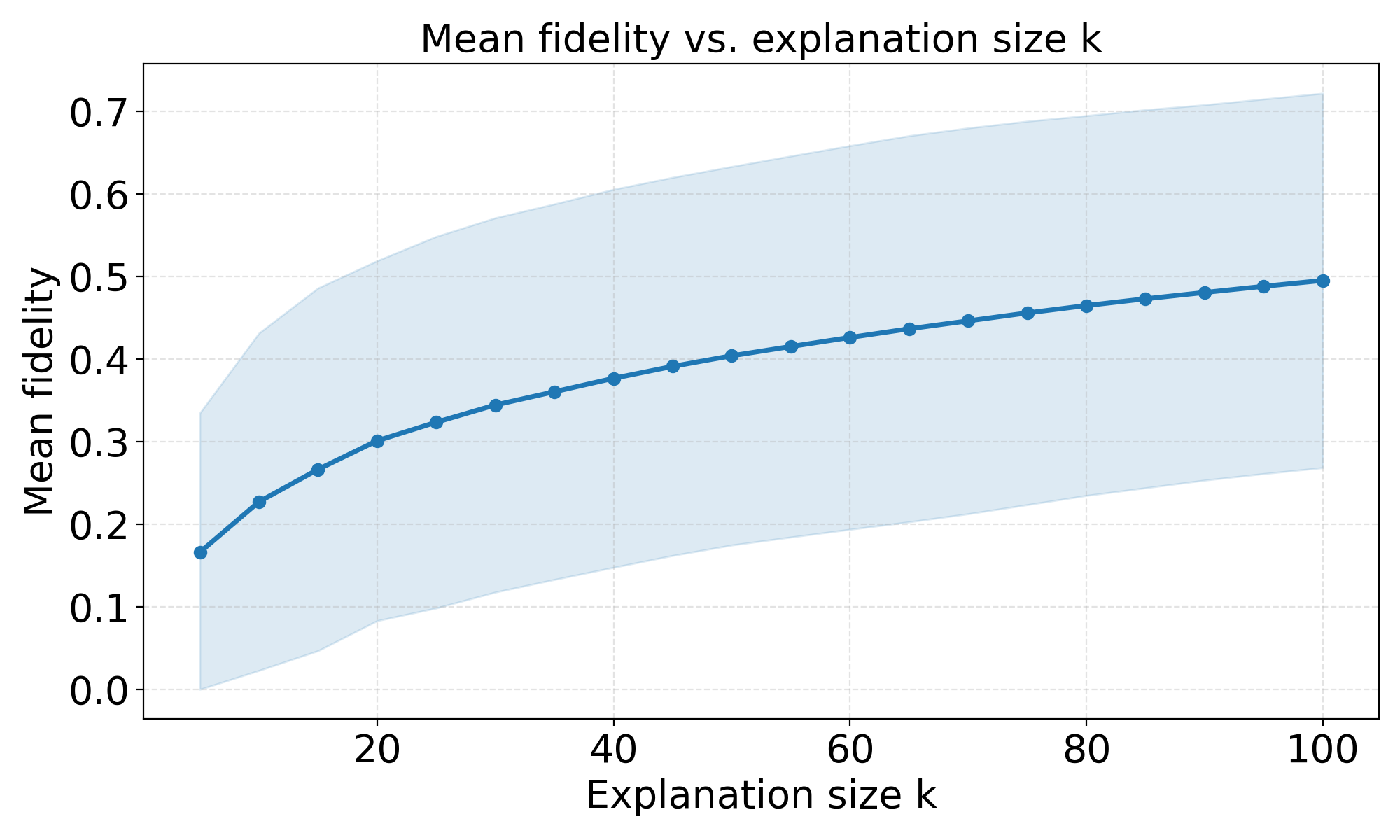}
    \caption{Fidelity score as a function of $k$}
    \label{fig:fidelity_vs_k}
\end{figure}

We ran xLightGCN on the Gowalla dataset (https://snap.stanford.edu/data/loc-gowalla.html), a large scale benchmark dataset for recommendation systems where users are recommended geographical locations (points of interest). We tested across a range of possible $k$ values to assess the relationship between $k$ and fidelity score (Figure~\ref{fig:fidelity_vs_k}), each averaged across 100 randomly chosen user/item pairs, choosing the $k$ range to be between 0 and 100, as any larger would be a dense subgraph difficult to interpret.

\begin{figure}[t]
    \centering
    \includegraphics[width=0.9\columnwidth]{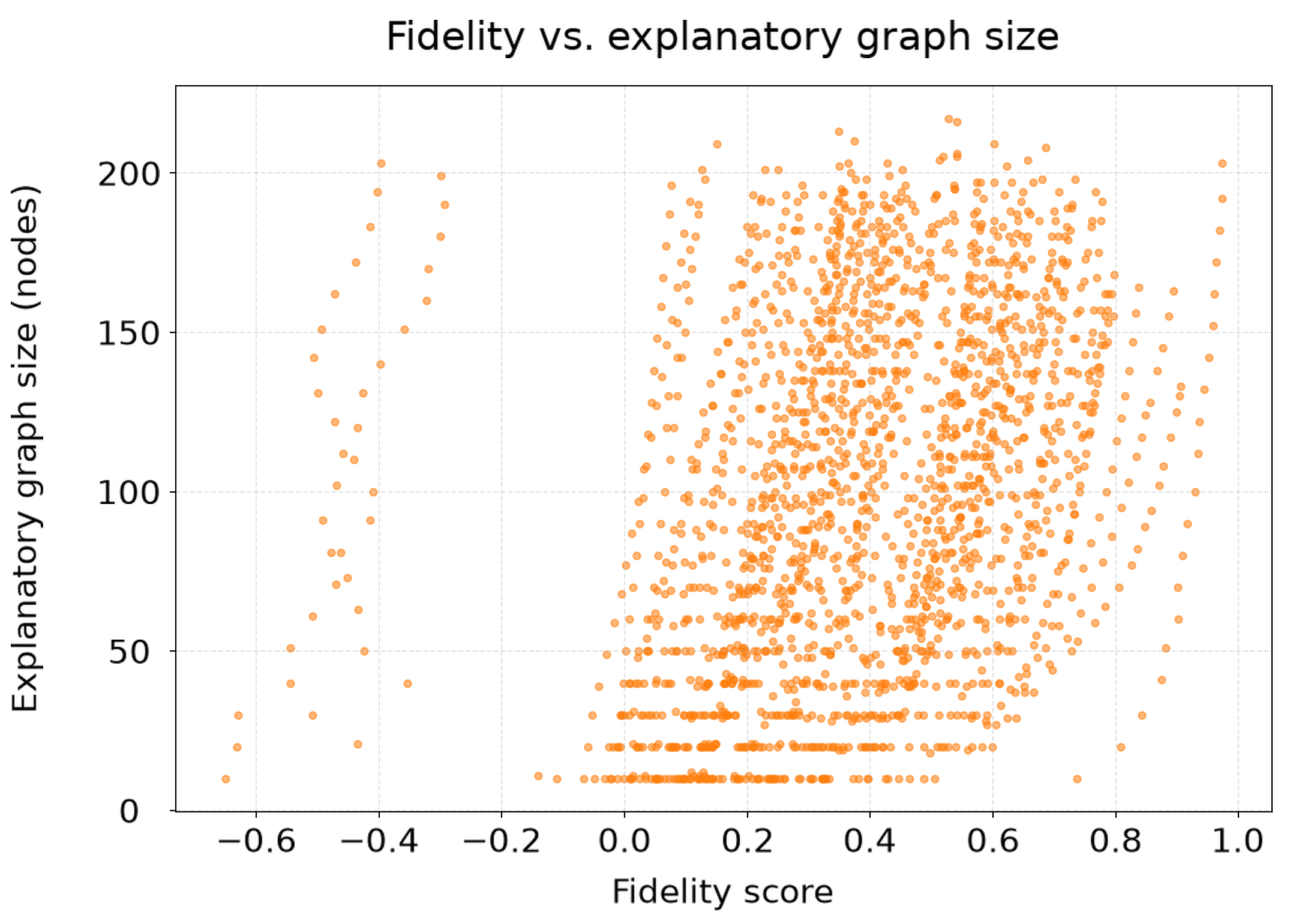}
    \caption{Explanatory subgraph size needed to reach a particular fidelity threshold.}
    \label{fig:fidelity_vs_graph_compactness}
\end{figure}

We also assess the size of the explanatory subgraph required to reach a certain fidelity threshold on average, (Figure~\ref{fig:fidelity_vs_graph_compactness}).

These results show a tradeoff between explanatory subgraph size $k$ and fidelity $F(u,i)$: Larger subgraphs explain more of the model's prediction but are more difficult to interpret due to their complexity. The relationship between fidelity and $k$ shows diminishing improvements as $k$ increases, suggesting that a few influential nodes are key to explaining most of the prediction, providing empirical support for the effectiveness of xLightGCN for isolating a small, interpretable subgraph that explains most of a recommendation system's prediction.

\section{Conclusion}

In this paper, we presented xLightGCN, a model-specific framework for generating compact explanations of recommendations produced by LightGCN family architectures. xLightGCN identifies the nodes that most strongly influence an individual recommendation and represents these contributions as an interpretable subgraph, providing evidence grounded in the actual mechanism producing the prediction. Furthermore, the unexplained residual and resulting fidelity score provide a quantitative measure of how completely the explanation accounts for the model's output. The explanations produced by xLightGCN together with a quantitative indication of the coverage of that explanation provide evidence that can support transparency and auditing, and can assist developers in evaluating the trustworthiness of their recommendation systems. The efficiency gains generated by our method also provide added-value.

Our work illustrates a broader role for explainability in trustworthy AI: explainability can serve as more than a user-facing interpretability feature; it can be an assurance mechanism for identifying, auditing, and quantifying the evidence underlying model outputs.

Several limitations remain. Most importantly, an explanation does not inherently mean that a system is trustworthy or accurate, it just provides a tool through which these properties can be investigated. In addition, xLightGCN relies on the linear propagation structure of LightGCN and is therefore not applicable directly to arbitrary GCN or recommendation architectures. This specialization is what enables its exact and efficient decomposition, suggesting that architecture-specific explainability can be valuable when strong assurances about model behavior are required.

Future work should investigate explainability frameworks in other architectures with tractable computational structures. Also, the paradigm of isolating influential nodes in order to understand complex network behavior should be explored further; the work done on Dynamic Opinion Networks \cite{muppasani2024planning}, \cite{muppasani2024expressive}, may provide further valuable insights. In addition, we argue that explainability offers strong insight in assessing the trustworthiness of AI systems, and more research needs to be done exploring its role in developing trustworthy, safe, and assured AI.

\bibliography{references}

\end{document}